\documentclass[runningheads]{llncs}
\usepackage{graphicx}
\usepackage{rotating}
\usepackage{floatpag}
\rotfloatpagestyle{empty}
\usepackage{floatrow}
\usepackage{url}
\usepackage{caption}
\usepackage{subcaption}
\usepackage[dvipsnames]{xcolor}
\usepackage{tabularx}
\usepackage{booktabs}
\usepackage{fancyhdr}
\usepackage{amsmath}

\begin{document}

\title{Towards Discourse Parsing-inspired\\ Semantic Storytelling}

\author{Georg Rehm\inst{1} \and
Karolina Zaczynska\inst{1} \and
Juli\'{a}n Moreno-Schneider\inst{1} \and \\
Malte Ostendorff\inst{1} \and 
Peter Bourgonje\inst{1} \and 
Maria Berger\inst{1} \and \\
Jens Rauenbusch\inst{2} \and 
Andr\'{e} Schmidt\inst{2} \and 
Mikka Wild\inst{2}}

\authorrunning{Georg Rehm et al.}

\institute{DFKI GmbH, Alt-Moabit 91c, 10559 Berlin, Germany 
\and
3pc GmbH Neue Kommunikation, Prinzessinnenstra\ss{}e 1, 10969 Berlin, Germany \\[1ex]
Corresponding Author: Georg Rehm -- \email{georg.rehm@dfki.de}}

\maketitle              

\begin{abstract}
Previous work of ours on Semantic Storytelling uses text analytics procedures including Named Entity Recognition and Event Detection. In this paper, we outline our longer-term vision on Semantic Storytelling and describe the current conceptual and technical approach. In the project that drives our research we develop AI-based technologies that are verified by partners from industry. One long-term goal is the development of an approach for Semantic Storytelling that has broad coverage and that is, furthermore, robust. We provide first results on experiments that involve discourse parsing, applied to a concrete use case, ``Explore the Neighbourhood!'', which is based on a semi-automatically collected data set with documents about noteworthy people in one of Berlin's districts. Though automatically obtaining annotations for coherence relations from plain text is a non-trivial challenge, our preliminary results are promising. We envision our approach to be combined with additional features (NER, coreference resolution, knowledge graphs). 
\end{abstract}

\thispagestyle{fancy}
\fancyhf{}
\fancyhead[R]{}
\fancyfoot[C]{\tiny Copyright \textcopyright\ 2020 for this paper by its authors.\\ Use permitted under Creative Commons License Attribution 4.0 International (CC BY 4.0).}
\renewcommand{\headrulewidth}{0pt}

\section{Introduction} 
\label{sec:introduction}

Cultural institutions such as museums, archives or libraries often rely on public funding and therefore need to communicate their value to the public constantly. One successful way to achieve this goal is to employ storytelling, which can be defined as creating emotional, interactive narratives in a digital format. Storytelling enables cultural institutions to make use of their digitized collections, demonstrating their relevance and reaching out to new audiences. Due to the extremely large amounts of available digital content, the curation of stories is typically performed by human knowledge workers. This calls for automated procedures. Such procedures should 1) label the content for several types of metadata semi-automatically, allowing for relevant categorisation. And 2) process the individual content pieces to present the information contained in them to a knowledge worker in an intuitive way. Since cultural organisations are often unlikely to be able to face this challenge on their own, we develop a platform supporting this use case in the the technology transfer project QURATOR. Our goal are semi-automatic technologies that keep the human in the loop and allow for fast, efficient and intuitive exploration of large and highly domain-specific data sets. Relating events into a schematic structure, i.\,e., storytelling, and ordering them, e.\,g., in terms of topic, locality or causal or temporal relationships, aid humans in finding meaningful patterns in data \cite{Bruner1991-BRUTNC}. 

In earlier work, we described approaches to Semantic Storytelling making use of Named Entity Recognition (NER) and Event Detection  \cite{moreno2017semantic,rehm2017h,rehm2018}. In this article, we explore ways to present a knowledge worker the semantic structure between text segments in an incoming text collection, making it possible to find interesting and surprising connections and information inside texts regarding a predefined topic. We focus on means of relating text segments to each other by borrowing from frameworks for the processing of coherence relations. From Rhetorical Structure Theory (RST) \cite{mannthomp88} we borrow the idea that larger sequences of texts (i.\,e., non-elementary discourse units) are related, moving beyond the shallow parsing of individual coherence relations.
From the Penn Discourse TreeBank (PDTB) \cite{Prasad08thepenn} we use the sense inventory and perform a series of experiments, relating text segments according to the four top-level classes of the PDTB sense hierarchy. The experiments are centered around the use case ``Explore the Neighbourhood!''. This tool, currently in development, is an urban exploration app that makes uses of documents on the Berlin district of Moabit. It allows users to follow stories, created by an editor semi-automatically, while exploring the district both physically and digitally.

The remainder of this paper is structured as follows. Section~\ref{sec:relatedwork} reviews relevant work, in particular, approaches using discourse relations in text. Section~\ref{sec:industry} explains the use case in more detail. Section~\ref{sec:sst} provides a technical definition, while Section~\ref{sec:experiments} outlines the experiments on the data set we created. Finally, Section~\ref{sec:conclucsions} provides a summary and suggests directions for future work.

\section{Related Work}
\label{sec:relatedwork}

The act of storytelling and the resulting stories, can be seen as a strategy to uncover meaningful patterns in the world around us \cite{Bruner1991-BRUTNC}. At the core of research on classical narratology, essential to storytelling, is the uncovering of the rules that underlie this strategy, or at least the ways to best achieve the goal. Early work on narratology is described in \cite{bal1985narratology}, defining a narrative as a discourse following a plot structure that has a chronological and logical event order. More recently, \cite{caselli2017} applied this definition of plot structure to (chrono)logically ordered events.
Another line of work on narratology is represented by the work of \cite{propp1968morphology}, who analyzes the basic, irreducible, structural elements of Russian folk tales. More recently, Propp's work was used by \cite{yarlott2016proppml} for their story detection and generation systems. The same authors, in \cite{yarlott2018identifying}, make use of another field of research related to text coherence, namely that of the processing of coherence relations. They apply the work of \cite{van2013news} on hierarchical discourse relations to work out how paragraphs behave when being used as discourse-structural units in news articles, with the ultimate goal of understanding the importance and temporal order of story items. Our work follows a similar approach, but uses PDTB sense hierarchy labels.
The PDTB \cite{Prasad08thepenn} is an (English) corpus of Wall Street Journal articles (a subsection of the Penn TreeBank \cite{Marcus1994}) annotated for individual discourse relations. We adopt the PDTB sense hierarchy, because it is the single largest corpus annotated for coherence relations and therefore the corpus best facilitating machine-learning based approaches. Due to the shallow nature of the PDTB framework (it only annotates individual relations, without making commitment to larger text structure, or mutual importance or relevance), we additionally source from RST \cite{mannthomp88}, particularly the notion of nuclearity. In RST, a text is divided into \emph{Elementary Discourse Units}, which are joined together, forming either a \emph{mono-nuclear} relation (with one unit being the more prominent, important or relevant \emph{nucleus} and the other, less prominent unit being the \emph{satellite}) or a \emph{multi-nuclear} relation. It is this notion of prominence, or relative importance to the storyline at hand, that we adopt from RST.

With regard to application-driven approaches, much work has been done on the final, surface realisation aspect of text generation \cite{fan2018hierarchical,fan2019strategies}. An approach resembling more closely ours is described by \cite{nie2019dissent}, who use dependency parsing in combination with discourse relations to determine sentence relations. In our approach, however, in addition to finding relevant articles for the user, we want to classify the type of relation the articles in question have to each other.

In our own previous work we described tools supporting the processing and generation of digital content with a strong industry focus, as is equally the case in the current context of the QURATOR project. The functionality of the curation technology platform is explained in \cite{rehm2018}. \cite{rehm2017h} presents an example of this platform applied to the use case of a personal communication archive, i.\,e., a collection of approx.~2,800 letters exchanged between the German architect Erich Mendelsohn and his wife Luise between 1910 and 1953. From this, we extracted, i.\,a., named entities, temporal expressions and events, combined these and used them to track and visualise the movement (across the globe) of Erich and Luise. Additional prototypes are presented in \cite{rehm2017c} and \cite{rehm2017b}.

\section{Industry Needs and Applications: The ``Explore the Neighbourhood!'' Use Case}
\label{sec:industry}

“Explore the Neighbourhood!” is a concept for a mobile app, which engages urban explorers in semi-automatically created stories, making use of digitized cultural collections. Moabit is a district in Berlin and was chosen due to its rich history and lively present. Such an app could be made available by museums, cities or municipalities, tourist information offices or local marketing campaigns. End users might be tourists, pupils studying in or visiting the neighborhood, or residents. Value is created for all parties by entertaining and educating users whilst communicating the district's or cultural institution’s relevance. The app offers both curated and generated stories. While in a final concept of “Explore the Neighbourhood!” these differences might not be noticed by the end user, in the following we will present each approach separately to describe the concept more precisely. We plan to fully integrate the approach described in Section~\ref{sec:sst}.

\subsection{Curated Stories}

\begin{figure}[!ht]
  \centering
  \includegraphics[width=1.0\textwidth]{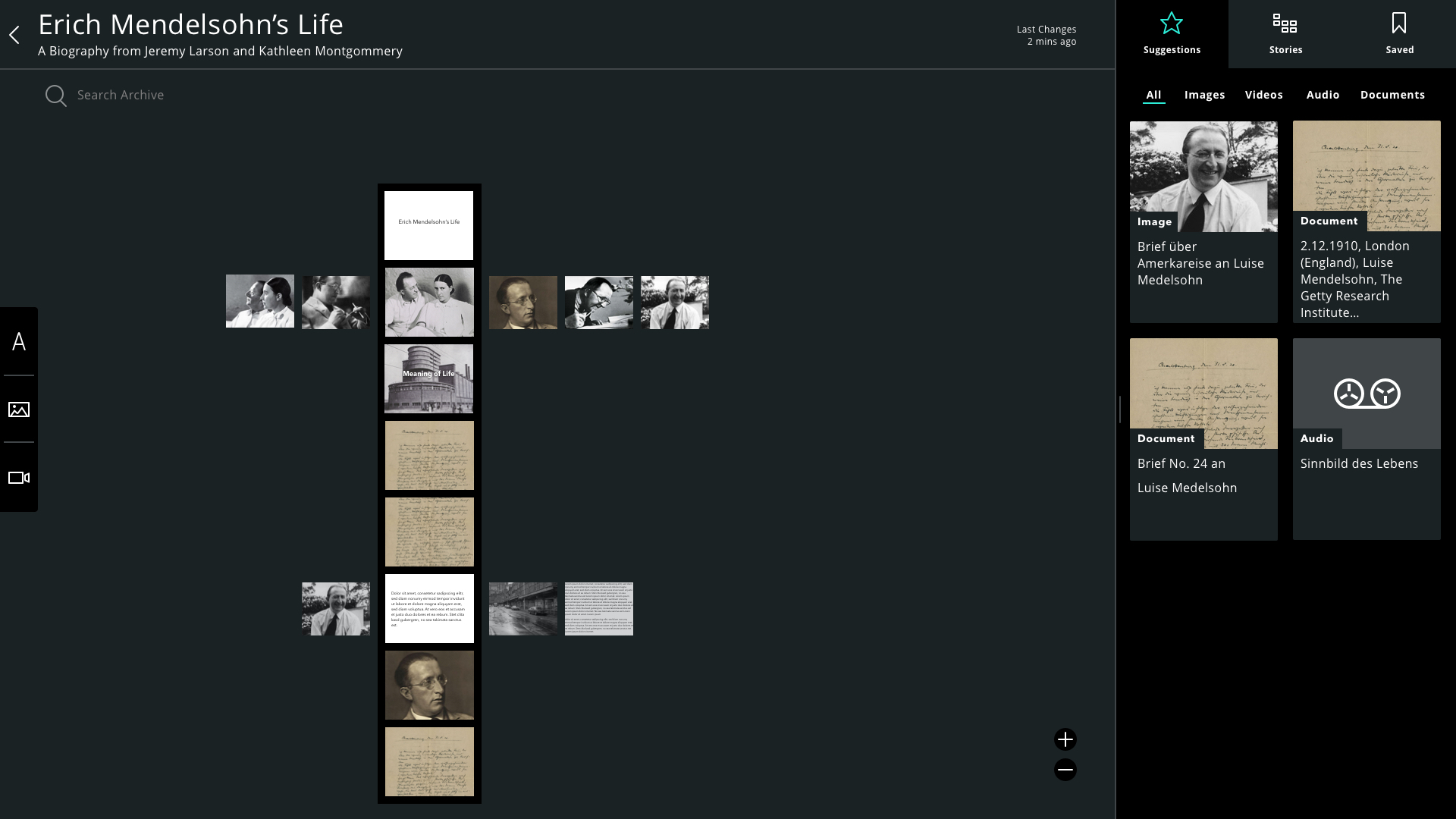}
  \caption{The smart
  authoring environment} \label{fig:fig5}
\end{figure}

Upon launching the app a set of interactive stories is offered to the end user who can influence the story’s direction, depth, and pace. Nevertheless, it still contains significant plot elements curated by an editor. The curation process requires the editor to define several storylines in a customised tool, which contains search capabilities and a recommendation system (Figure~\ref{fig:fig5}), both of which help surface relevant content for each step along a story path. Such a tool is made possible due to rich metadata which allow queries such as “poems describing Berlin in a praising tone” (text classification and analysis detecting locations and sentiment) or “photos showing Kurt Tucholsky next to a church” (image classification and analysis detecting people and objects, in this case churches). Figure~\ref{fig:fig5} shows the user interface of such a tool.

\begin{figure}
    \centering
    \begin{subfigure}[b]{0.4\textwidth}
    \centering
        \includegraphics[width=\textwidth]{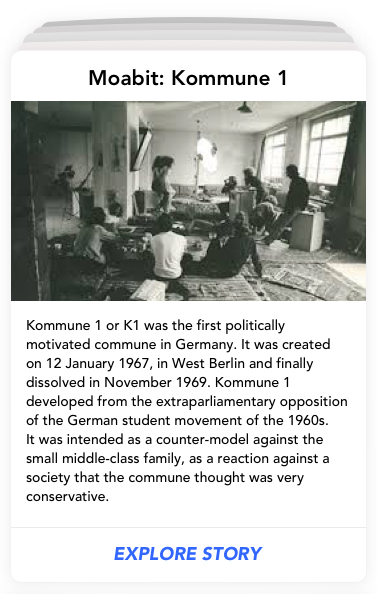}
        \caption{Description of Kommune 1}
        \label{fig:3pcfig1}
    \end{subfigure}
    ~ 
    \begin{subfigure}[b]{0.4\textwidth}
    \centering
        \includegraphics[width=0.75\textwidth]{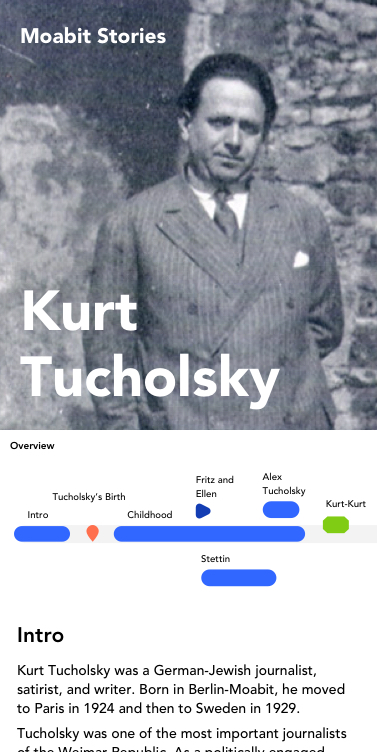}
        \caption{Kurt Tucholsky's biography}
        \label{fig:3pcfig2}
    \end{subfigure}
    ~ 
    \caption{The app allows the exploration of different aspects of Moabit}\label{fig:figs12}
\end{figure}

Curated stories can be published to the app  (Figure~\ref{fig:3pcfig1}). Stories may contain geographical points of interest within Moabit which are connected through an overall story arch, such as a biography. The exemplary stories depicted in this article follow the biography of Kurt Tucholsky (Figure~\ref{fig:3pcfig2}), a German-Jewish journalist and writer born in Moabit in 1890. The stories contain locations, historic photos and maps, scanned original works and editorial content.

The existence of several storylines within a story, as well as several stories in parallel, allows for connections to be forged. These connections can be based on common topics, locations, or other parameters that support a consistent and emotional narrative. Users can follow one path through a story, choose to dive deeper into certain aspects of it, e.\,g., Kurt Tucholsky (Figure~\ref{fig:3pcfig2}), change their perspective onto a topic by exploring alternative stories, or switch to a completely different, yet connected story. The consumable stories are linked in a network and limited only by the amount of pieces of information and the size of the network created by the editor, who can extend it continuously.


\subsection{Generated Stories}

Unlike curated ones, generated stories are created entirely by a storytelling engine. This is made possible due to a set of well-chosen parameters which influence the automatic selection and connection of content. These parameters are defined by several factors:

\begin{itemize}
\item a chosen topic (initiated through a keyword or phrase)
\item the type of story being told (such as biography or travel guide),
\item users’ preferences (such as available time, current sentiment, preferred mode of travel),
\item users’ behavior (such as current location, walking speed, orientation).
\end{itemize}

Based upon the factors listed above, “Explore the Neighbourhood!” automatically generates a story by selecting the right content based on its rich metadata. The end result, which is the story consumed by the users, may not look so different from editor-curated stories. Nevertheless, since generating a story happens in real-time, it constantly adapts to users’ choices, which creates a more personal and more interactive experience.


\section{Semantic Storytelling: Technical Description}
\label{sec:sst}

One of the goals of our Semantic Storytelling system is to aid knowledge workers in selecting relevant pieces of content, e.\,g., the app editor who wants to curate stories for the app. Following the prototype of the ``Explore the Neighbourhood!'' app (Section~\ref{sec:industry}), this section describes the technical details of the back-end. 

\begin{figure}[htbp]
\centering
\includegraphics[width=\textwidth]{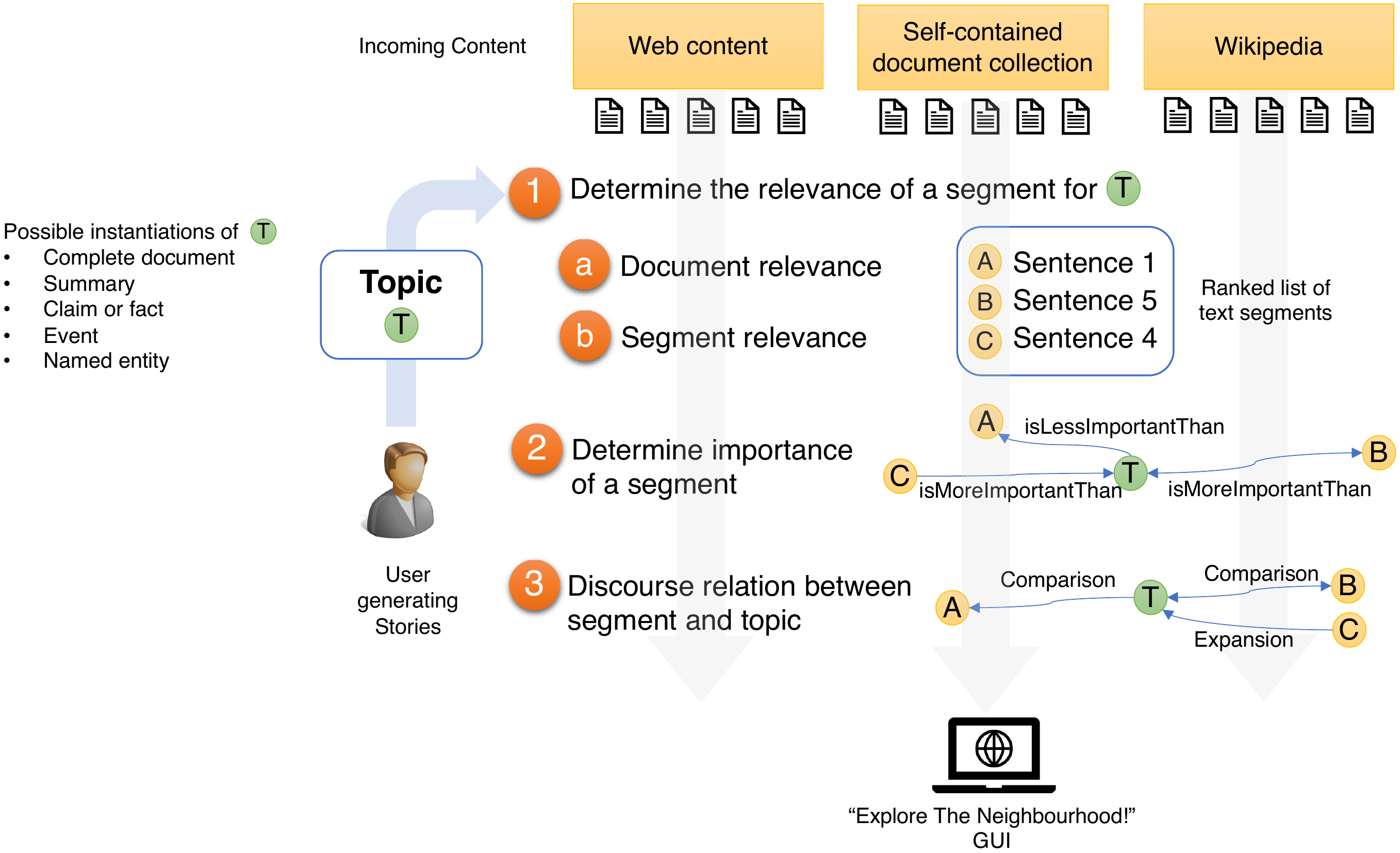}
\caption{Architecture of the Semantic Storytelling approach} 
\label{fig:approacharchitecture}
\end{figure}

Let us assume the following situation. A user is visiting a city and wants information about a topic $T$ regarding the neighbourhood. The goal of the curation prototype is, then, to identify and to suggest new content for the app that can be included in the user's tour. To do so, we first have to initialise the topic $T$, for example, as a sentence,  keyword or named entity. Next up, the tool has to identify if, for example, a document in a collection or a web page is \emph{relevant} for topic $T$, and, if so, if it is \emph{important} for $T$. Finally, we identify the \emph{semantic relation} between incoming texts and the provided topic $T$, which could be, among others, \emph{background}, \emph{cause}, \emph{contrast}, \emph{example} etc. In the following, we describe these steps in more detail (Figure~\ref{fig:approacharchitecture}). 

\subsubsection{Step 1: Determine the Relevance of a Segment for a Topic}
The approach starts with a topic $T$, instantiated through a text segment such as a complete document, a headline or a named entity. To identify content pieces relevant for $T$, we process incoming textual content, like a self-contained document collection, a systematically compiled corpus or a knowledge base. 

For each piece of content, we need to decide whether its topic is relevant for $T$, which can be computed in various ways. We can employ topic modeling (LDA, LSA) or, without explicitly modeling topics, we can also perform pair-wise comparisons of document similarity. Document pairs with a high similarity score are assumed to cover the same topic, therefore, we start with the seed document $d_s$ of which we know that it represents $T$ and measure its similarity to other candidates. To compute semantic similarity, documents are represented as numerical vectors. Classical methods like bag-of-words or tf-idf encode documents as sparse vectors \cite{salton1983}, while neural methods (word2vec, sent2vec, doc2vec, see e.\,g., \cite{mikolov2013efficient,pagliardini2017unsupervised,selivanov2016text2vec}) produce dense representations. In both cases, cosine similarity can be used to compute the similarity of the document vectors. 

\subsubsection{Step 2: Determine the Importance of a Segment}

If we have determined all documents $d$ which are related to $T$, we need to determine the importance of $d$ (or its segments or sentences) with regard to $T$. There is no off-the-shelf approach to determine the importance of a segment with regard to a topic, but various cues and indicators can potentially be exploited. One way of doing this is to borrow from RST, especially the notion of nuclearity. Constructing an RST tree involves decisions with regard to the status of text segments including their discourse relation to other segments and also regarding their role as a \emph{nucleus} (the important core part of a relation) or \emph{satellite} (the contributing part of a relation) in the context of a specific discourse relation. Two segments are assigned either a satellite-nucleus (S-N), nucleus-satellite (N-S) or a nucleus-nucleus (N-N) structure. This sub-task can be done in isolation \cite{journals/dad/HernaultPdI10,soricut-marcu-2003-sentence}, or in conjunction with the relation classification task \cite{jotycodra}. When performed iteratively, this pair-wise classification can result in a set of most important segments regarding $T$. Another way of determining topical importance is to treat it as a segment-level question answering task. Given a document $d$ consisting of text segments $(t_1, t_2, \dots t_n)$, the aim is to find the segment $t_i$ that contains the answer to the input question (i.\,e., topic $T$). Transformer language models have achieved state-of-the-art results for question answering \cite{Devlin2018}, suggesting that those model architectures would be beneficial for storytelling.

\subsubsection{Step 3: Semantic or Discourse Relation between two Segments}

After having established the relevance and relative importance, we proceed with determining the semantic or discourse relation that exists between the text segments and topic $T$. Our initial experiments are based on the PDTB due to its considerably larger size with more than 1.1 million tokens over the RST-Discourse TreeBank \cite{Carlson-corpus:2002} with approx.~200k tokens. We adopt the PDTB's sense hierarchy, which comprises four top-level classes, 16 types on the second level and 23 sub-types on the third. For now, our experiments are based on the top-level senses, \emph{Temporal}, \emph{Contingency}, \emph{Comparison}, \emph{Expansions}, and an additional label, \emph{None}.

\section{Experiment for ``Explore the Neighbourhood!''}
\label{sec:experiments}

In this section, we describe our first experiments, which aim to explore the suitability of the approach and helps us gain an understanding of what we can achieve in the long run. We concentrate on step 3, therefore, we created a data set of crawled web documents about the Berlin district Moabit, and implemented initial experiments to classify discourse relations between text segments inside the data set. We would like to show a comparison with similar tools, but to the best of our knowledge, there are no similar tools that are extracting semantic relations through intra-document text segments (using PDTB).

\subsection{Data Set}
\label{sec:data set}

The data set is composed of documents containing information and stories connected to the district of Moabit in Berlin. We are in the first stages of developing this data set. In the long term, the idea is to put together a much larger collection of documents focused on Moabit so that it can be used for the Semantic Storytelling prototype. We used the focused crawler Spidey\footnote{\url{https://github.com/vikrambajaj22/Spidey-Focused-Web-Crawler}}, which returns a list of URLs from websites which are based on a set of predefined query terms. We manually defined 28 queries about interesting places, buildings, or persons connected to Moabit. Some of these terms are \emph{Moabit}, \emph{Moabit gentrification}, \emph{Kleiner Tiergarten}, \emph{Kulturfabrik Moabit}, \emph{Berlin Central Station} and \emph{Kurt Tucholsky}. After obtaining the website URLs, we crawl and boilerplate the content of the pages and their metadata\footnote{We use Newspaper3k, see~\url{https://github.com/codelucas/newspaper}}. The resulting data set is composed of slightly more than 100 documents that have been filtered manually in a second step. 

\subsection{Classifiers for Discourse Relation between Text Segments}

Our aim is to extract discourse relations from texts and so, being able to extract relevant content from a text collection and, in the longer run, to find new storylines composed of semantically related parts of different text segments taken from the collection. We train a relation sense classifier on PDTB2 \cite{Prasad08thepenn} and apply it on two pieces of content. For training, we use the two arguments of a relation, but at a later point we deploy it using individual sentences. We argue that the sentence-level is the most appropriate level to use as input for our classifier (as opposed to the shorter token or phrase level, or the longer paragraph level) and that the discrepancy between argument shapes and typical sentence lengths (itself very much dependent on the domain) is tolerable.

\subsubsection{Classifier Model}

Classifying the discourse relation between sentence pairs requires a semantic understanding of the sentences. We encode the text as deep contextual representations with a language model based on the Transformer architecture \cite{Vaswani2017}. To be precise, the pre-trained language model from DistilBERT \cite{Sanh2019}, a distilled version of Bidirectional Encoder Representations from Transformers \cite{Devlin2018} is used\footnote{We use the PyTorch implementation by HuggingFace \cite{Wolf2019}.}. BERT performs well for document classification tasks \cite{Ostendorff2019}. 

\begin{figure}[!ht]
\floatbox[{\capbeside\thisfloatsetup{capbesideposition={right,center},capbesidewidth=0.5\textwidth}}]{figure}[\FBwidth]
{\caption{The architecture of the Siamese BERT model for the classification of discourse relations between two text segments $d_1$ and $d_2$. The output of the classification layer $\hat{y}$ holds the predicted semantic relation according to the top-level PDTB2 senses.}\label{fig:test}}
{\includegraphics[width=0.43\textwidth]{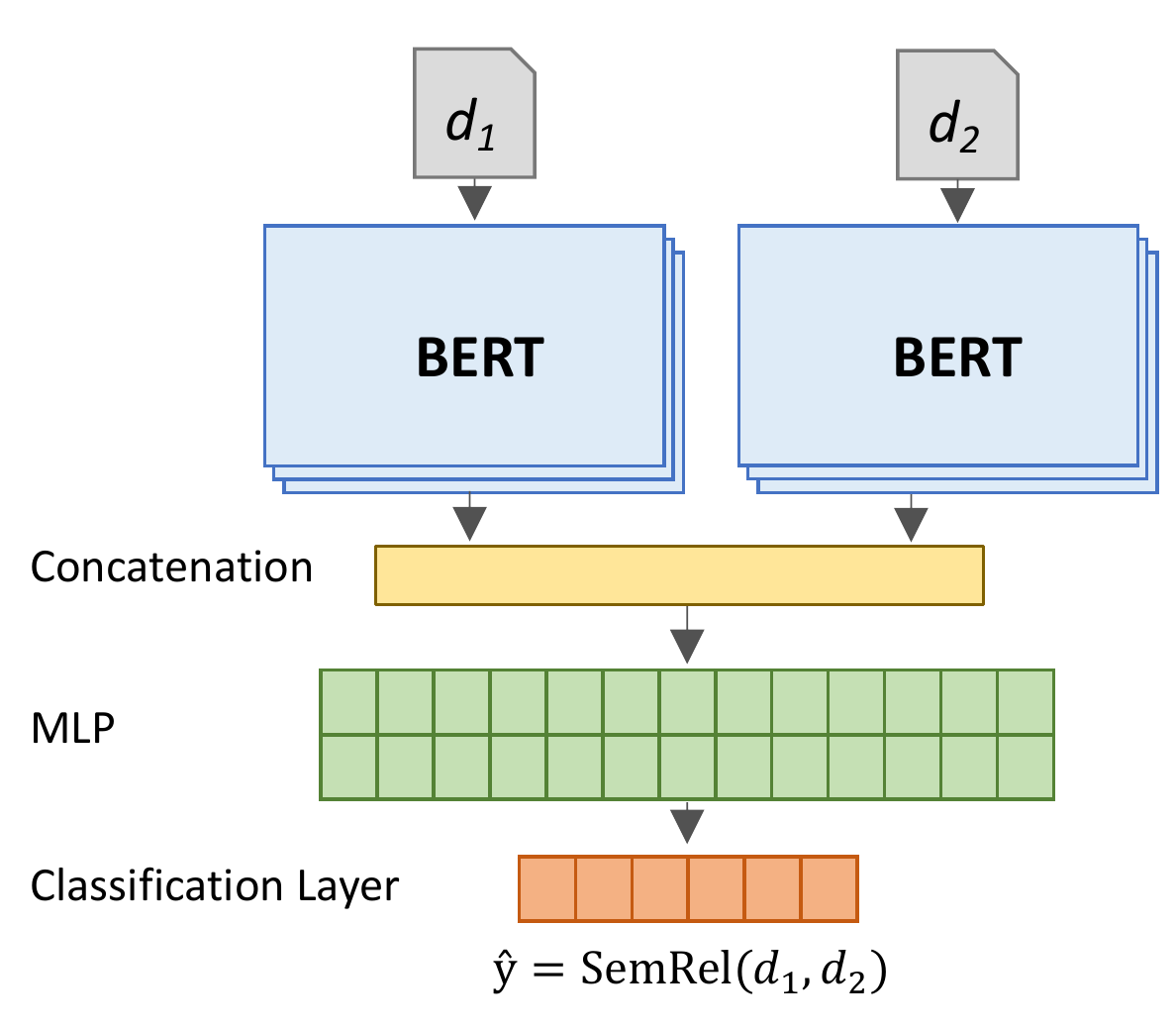}}
\end{figure}

To classify the relation between two texts, we employ a Siamese architecture \cite{Bromley1993}. In contrast to a classical Siamese model, in which a binary classifier is employed on the output of the two identical sub-networks, we feed the sub-network output into a multi-label classifier, as illustrated in Figure~\ref{fig:test}.  

Text snippets $d_1$ and $d_2$ are inputs to the classifier. BERT's architecture consists of six hidden layers, each layer consists of 768 units (66M parameters; DistilBERT). BERT is used in a Siamese fashion such that $h_i=\text{BERT}(d_i)$ is the encoded representation of text $d_i$ where $h_i$ is the last hidden state of the last BERT layer. The final feature vector $x_f$ is a combined concatenation of the text representations:

\begin{equation}
x_f=[h_1; h_2; |h_1-h_2|; h_1 * h_2; \frac{h_1+h_2}{2}]
\end{equation}

\noindent On top of the concatenation, we implement a Multi-Layer Perceptron (MLP). The MLP consists of two fully-connected layers, $\mathcal{F}_f(\cdot)$ and $\mathcal{F}_g(\cdot)$, where each layer has 100 units and $\text{ReLU}(\cdot)$ is the activation function. The discourse relation $\hat{y}$ is classified on the basis of the feature vector $x_f$ as follows:

\begin{equation}
\hat{y}=\sigma(\mathcal{F}_f(\text{ReLU}(\mathcal{F}_d(x_f))))
\end{equation}

\noindent The logistic softmax function $\sigma(\cdot)$ generates probabilistic multi-label classifications. The dimension of $\hat{y}$ corresponds to the number of classification labels, which are the four top-level PDTB2 senses (\emph{Temporal}, \emph{Contingency}, \emph{Comparison}, \emph{Expansions}) and one additional dimension (\emph{None}).

\subsubsection{Transfer Learning}

Our target corpus of texts for ``Explore the Neighbourhood!'' does not include any kind of annotated training data. Thus, we cannot use the data set to train the classifier. Instead, we rely on the PDTB2 data set. Training is performed with batch size $b=16$, dropout probability $d=0.1$, learning rate $\eta=2^{-5}$ (Adam optimizer) and 5 training epochs. These hyperparameters are the ones proposed by \cite{Devlin2018} for BERT fine-tuning.

\begin{table}[h!]
\centering
\footnotesize
\begin{tabular}{ l c c c c} \toprule
  \textbf{PDTB Relation}     & \textbf{~~Precision~~} & \textbf{~~Recall~~} & \textbf{~~F1-score~~} & \textbf{Support} \\ \midrule
 \emph{Comparison}     & 0.50      &   0.47 &  0.48    &   1598  \\
 \emph{Contingency}    & 0.38      &   0.65 &  0.48    &   1582  \\
 \emph{Expansion}      & 0.50      &\textbf{0.79}&\textbf{0.61}& 2993  \\
 \emph{Temporal}       &\textbf{0.51}& 0.55 &  0.53    &    869  \\
 \emph{None}           & 0.49      &   0.73 &  0.59    &   1078  \\ \midrule
 Micro avg.      & 0.47      &   0.67 &  0.55    &   8120  \\
 Macro avg.      & 0.48      &   0.64 &  0.54    &   8120  \\ \bottomrule
\end{tabular}
\caption{Results of training multi-class prediction based on PDTB2 data set in a 80-20 train-test-split.}
\label{tab_classifier_acc}
\end{table}

The results that are derived from a 80-20 train-test-split are shown in Table~\ref{tab_classifier_acc}. For evaluation, we use the multi class metric F1-micro average, which calculates the metrics globally by counting the total true positives, false negatives and false positives to compute the average metric. In a multi-class classification setup, micro-average is preferable if you suspect there might be class imbalance. In the end, we achieve 0.55 micro average F1. Due to the fact that we have not implemented features relating to the connective, our classification performs lower than current state-of-the-art approaches.

\subsection{First Experiment on Use Case Data Set and Discussion}

Given the PTDB2-based classifier, we continue to find discourse relations within the corpus containing documents for the ``Explore the Neighbourhood!'' use case. As a preprocessing step, we first exclude all non-English documents and group documents by topic based on the query terms for the focused crawler. Next, we find document pairs among the topic groups (only semantically similar document pairs are considered). More precisely, documents are represented as tf-idf vectors and the cosine similarity of a document pair $d_a$ and $d_b$ must be above a fixed threshold ($cosine(d_a,d_b) >0.15$). Our classifier is trained to detect sentence-level relations, thus, we also split the documents into sentences\footnote{We use pySBD, see \url{https://github.com/nipunsadvilkar/pySBD}}. After excluding all sentences with less than five words, we end up with 96,796 sentence pairs that are passed to the classifier. 

\begin{table}[htbp]
\scriptsize
\begin{tabular}{lp{1.5cm}p{3.25cm}p{3.25cm}cccccc}
    \toprule\addlinespace
    & \multicolumn{4}{c}{\textbf{Documents}} & \multicolumn{5}{c}{\textbf{Discourse relations}} \\
    \addlinespace\midrule\addlinespace
    & \multicolumn{1}{c}{\textbf{Topic}} & \multicolumn{1}{c}{\textbf{Segment A}} & \multicolumn{1}{c}{\textbf{Segment B}} & \textbf{S.} & \textbf{Co.} & \textbf{Ct.} & \textbf{E.} & \textbf{T.} & \textbf{N.} \\
    \addlinespace\midrule\addlinespace
    1 & \textbf{Farin Urlaub Moabit} & In April 2012 they released another album ``auch” (``also”) & At the age of 16, Vetter went on a school trip to London, and returned home as a punk with dyed blonde hair. & .51 &.01 & .01 & .04 & \textbf{.93} & .1 \\
    \addlinespace\midrule\addlinespace
    2 & \textbf{Uschi Obermaier} & In 1968 and 1969 Obermaier starred in Rudolf Thome's first two feature films, ``Detektive" and ``Rote Sonne" (``Red Sun"). & She played maracas in the band Amon Düül, aka Amon Düül I, on two albums: Collapsing (1970, released by Metronome) and Disaster (1972, released by BASF Records [de]). & .39 & .0 & .0 & \textbf{.96} & .03 & .02\\
    \addlinespace\midrule\addlinespace
    3 & \textbf{AEG turbine factory} & It is an influential and well-known example of industrial architecture. & However, when it came to AEG’s public image and public perception, the focus remained on Peter Behrens: the famous artist-cum-architect overshadowed the engineer. & .32 & \textbf{.62} & .16 & .15 & .01 & .06 \\
    \addlinespace\midrule\addlinespace
    4 & \textbf{Kurt Tucholsky Moabit} & Admittedly, Tucholsky is seldom recognized as a philosopher. & He saw himself as a left-wing democrat and pacifist and warned against anti-democratic tendencies -- above all in politics, the military and justice -- and the threat of National Socialism. & .25 & \textbf{.45} & .28 & .19 & .01 & .08 \\ 
    \addlinespace\midrule\addlinespace
    5 & \textbf{Schult-heiss Brewery} & ''Good people drink good beer,'' Hunter S Thompson once said, writing about a beverage that is considered to be typically German and is, of course, also popular in Berlin. & Schultheiss is currently brewing far less beer than at the time of re-unification. & .16 & .32 & .19 & \textbf{.42} & .01 & .06 \\ 
    \addlinespace\bottomrule
\end{tabular}%
\caption{Manually evaluated examples. On the right, the table shows the similarity score (S.) between sentences and the prediction scores for each discourse relation (Co.=Comparison, Ct.=Contingency, E.=Expansion, T.=Temporal, N.=None)}
\label{tab:examples}
\end{table}

To get a first impression on the applicability of our approach, and to motivate our next steps, we manually select five example sentence pairs to evaluate them qualitatively. The first line of Table~\ref{tab:examples} shows an example where the classifier correctly labels the discourse relation as \emph{Temporal}, most likely because of the temporal markers included. In the second line, the approach correctly identifies the discourse relation as an \emph{Expansion}, i.\,e., segment B can be seen as an extension of the biography described in segment A. Nevertheless, in other examples, the approach is often unable to handle coreference. The classifier is often not detecting a discourse relation between two segments, even if those segments reference the same entity, while one segment uses a pronoun for the entity. By implementing a preprocessing step with rudimentary coreference resolution we expect the classification to improve significantly. The classifier predicts the label \emph{Comparison} often when specific lexical markers, such as \emph{however}, \emph{but} or \emph{while}, appear in segment B, like in example 3. Example 4 is an exception, where the classifier predicts the relation \emph{Comparison} correctly without needing a lexical marker, but, generally we observe that this dependency on lexical features leads to wrong predictions. We see one reason in the fact that the sentences are taken from different sources, and the lexical markers for the discourse relation are therefore often missing, also even if semantically it can be seen as a \textit{Comparison}. This is the case in example 5, which is wrongly predicted as an \textit{Extension} while we interpret it as a \textit{Comparison} (paraphrased as 'Even if he is recognized as a philosopher, he saw himself as a political activist'). On the other hand, in other examples, the lexical markers cause false positives errors. Hence, future work will extent the number of preprocessing steps to better group text segments which have the same content and talk about the same entities, events or topics. 

\section{Conclusions}
\label{sec:conclucsions}

We describe first experiments in order to apply our Semantic Storytelling approach to an industrial use case. This use case, ``Explore the Neighbourhood!'', makes it possible to interactively create a city guide with adjusting interesting stories about a particular district built upon user-dependent parameters, such as predefined topics, keywords, etc. The basic idea is to automate storytelling by detecting discourse relations between texts segments of different sources on the same topic, which makes it possible to be able to detect and create new storylines extracted from a document collection. We describe the different steps in order to create a corresponding processing framework. In the experiment presented here, we focus on the third step of our approach, the classification of discourse relations between segments. By focusing more on steps one and two as described in Section~\ref{sec:sst}, we will be able to improve the results in further experiments. For example, we expect the classification to improve significantly by using coreference resolution during preprocessing. One way of improving the coreference resolution would be to pretrain the classifier on the coreference task first \cite{joshi-etal-2019-bert}. As data sets are still limited, we will expand the data set for our needs and create, in the longer run, annotations to develop a gold standard. 

\section*{Acknowledgements}

The research presented in this article is funded by the German Federal Ministry of Education and Research (BMBF) through the project QURATOR (Unternehmen Region, Wachstumskern, no.~03WKDA1A). \url{http://qurator.ai}

\bibliographystyle{./splncs04}
\bibliography{./bibliography}

\end{document}